\documentclass[10pt]{llncs}

\usepackage{cite} 
\usepackage{url}  
\usepackage[utf8]{inputenc} 
\usepackage{hyperref}

\usepackage{multirow}
\usepackage{rotating}
\newcommand{\Alo}{Colon}
\newcommand{\Gol}{Leukemia}
\newcommand{\Kha}{SRBCT}
\newcommand{\Sin}{Prostate}

\begin{document}
\title{Robustness of Random Forest-based gene selection methods}

\author{Miron B Kursa}
\institute{Interdisciplinary Centre for Mathematical and Computational Modelling,\\
     University of Warsaw\\
\email{M.Kursa@icm.edu.pl}
     }

\maketitle

\begin{abstract}
Gene selection is an important part of microarray data analysis because it provides information that can lead to a better mechanistic understanding of an investigated phenomenon.
At the same time, gene selection is very difficult because of the noisy nature of microarray data.
As a consequence, gene selection is often performed with machine learning methods.
The Random Forest method is particularly well suited for this purpose.
In this work, four state-of-the-art Random Forest-based feature selection methods were compared in a gene selection context.
The analysis focused on the stability of selection because, although it is necessary for determining the significance of results, it is often ignored in similar studies.

The comparison of post-selection accuracy in the validation of Random Forest classifiers revealed that all investigated methods were equivalent in this context.
However, the methods substantially differed with respect to the number of selected genes and the stability of selection.
Of the analysed methods, the Boruta algorithm predicted the most genes as potentially important.

The post-selection classifier error rate, which is a frequently used measure, was found to be a potentially deceptive measure of gene selection quality.
When the number of consistently selected genes was considered, the Boruta algorithm was clearly the best.
Although it was also the most computationally intensive method, the Boruta algorithm’s computational demands could be reduced to levels comparable to those of other algorithms by replacing the Random Forest importance with a comparable measure from Random Ferns (a similar but simplified classifier).
Despite their design assumptions, the minimal-optimal selection methods, were found to select a high fraction of false positives.
\end{abstract}

\section{Background}
DNA microarrays, with their ability to capture a substantial fraction of a cell state, are one of the most powerful tools in the molecular biology.
From a machine learning point of view, standard microarray experiments generate an information system in which each object (measurement) is described by a vector of features corresponding to expression levels of a large number of genes (often approaching full set of the identified genes for a certain organism).
Additionally, microarray experiments generate a decision corresponding to the investigated state, such as the presence of a disease, the application of a certain stimulation, the state of the organism, the tissue, etc.

Because the number of investigated genes is always much larger than the number of measurements in a DNA microarray experiment, gene selection with these data belongs to the $p \gg n$-class of problems, which is known to promote a number of issues related to the stability, statistical power and feasibility of certain methods.
Moreover, because a measured set of genes is almost always not specifically targeted for a certain decision (in the machine learning sense), these data will contain a large number of redundant features.

For these reasons, it is usally desired to reduce the dimensionality of a microarray dataset.
Dimension reduction is often achieved by feature selection (i.e., the removal of unnecessary features) because it is the only method that maintains a direct relationship between a feature and a gene \cite{Saeys2007}; this is why this process is often called \textit{gene selection} in the context of microarray data.

It is often assumed that gene selection both provides meaningful insight into the data (e.g., by providing a list of genes relevant to the investigated condition) and serves as a pre-processing step that optimises next methods in the analysis pipeline.

However, this assumption is wrong \cite{Nielsen2007} and fature selection may only have one of two aims that require different approaches and tools: finding the \textit{minimal optimal} subset of features that is the smallest that will allow a given classifier to achieve maximal accuracy, or fiding the \textit{all relevant} subset, that is of all features relevant to the analysed phenomenon.

This is because the goal of the minimal optimal selection is to optimise certain classifier, thus it will be affected by inherent biases of that method.
For example, it may favour genes with expression levels that have certain characteristics, like follow a specific distribution.
Also, in $p\gg n$ datasets, false associations that are equal to or stronger than the true association are very likely to arise at random.
While minimal optimal selection will greedily reduce blocks of redundant features, such artefacts that can displace relevant genes from the final selection and lower the stability and recall of the method.

Unfortunately, only the minimal optimal problem is traditionally tackled because both its application and assessment (in terms of post-selection accuracy) are straightforward.
Yet only the solution to the all relevant problem can enable deeper insight in mechanics of an analysed phenomenon that go beyond just identifying the brightest signs of its occurrence.

The Random Forest algorithm is popular in the life sciences because it supports $p\gg n$ datasets, is robust to large amounts of noise, requires little parameter tuning and requires no predictor transformation \cite{Touw2012,Diaz-Uriarte2006,Breiman2001,Cutler2012}.
Random Forest also natively produces a feature-importance measure that directly expresses the role of a feature in all interactions utilised in the model, including weak and multivariate ones. These characteristics make Random Forest a promising classification algorithm for gene selection tasks \cite{Diaz-Uriarte2006}.

To this end, a number of Random Forest-based feature selection methods have been proposed for gene selection.
In this work, four state-of-the-art Random Forest classifiers are analysed: the Artificial Contrasts with Ensembles (RF-ACE) \cite{Tuv2009,rface} and Boruta \cite{Kursa2010c} methods, which are all relevant approaches, and the Recursive Feature Elimination (RFE) and Regularised Random Forest (RRF) \cite{Deng2012a} methods, which are minimal-optimal approaches.

Whenever possible, methods were re-evaluated with all three feature importance measures provided by the Random Forest algorithm as well as the importance scores provided by the Random Ferns \cite{Kursa2012a} algorithm, which is similar to a Random Forest but relies on a simpler and more stochastic base classifier.

Because all machine learning algorithms are heuristic methods, the correctness and optimality of their solutions cannot be guaranteed.
Consequently, any methodology implementing these approaches must properly validate the results.
In particular, if only a single application of a machine learning algorithm is applied to an entire dataset, subtle errors with very serious consequences may be introduced \cite{Kursa2011,Ambroise2002}.
To avoid this limitation, the work presented here employed bootstrap \cite{Efron1993}, method where each selection procedure was replicated over 30 times by resampling the original dataset.
Moreover, apart from performing the usual analysis of post-selection classification accuracy, a novel self-consistency-based approach for assessing the stability and robustness of a gene selection method was developed and applied.

Because the sole aim of this work was to investigate the characteristics of various gene selection methods, all tests were performed on four standard pre-processed microarray datasets: \Alo{}, \Gol{}, \Kha{} and \Sin{}. Moreover, for clarity, no additional sources of information about the datasets, such as temporal context, gene ontology or microarray calibration techniques (e.g., RNA spike-ins) \cite{Yang2006} were considered.

\section*{Results and Discussion}
\subsection*{Post-selection classification accuracy}
The most common method for the assessment and tuning of feature selection methods is to perform an error analysis on a classifier trained on a set containing only the selected features.
This method is motivated by the seemingly obvious assumption that because the presence of noise and redundant features decrease classification accuracy, minimal error will be achieved with a set lacking these artefacts.

\begin{figure}
\centering
\includegraphics[width=\textwidth]{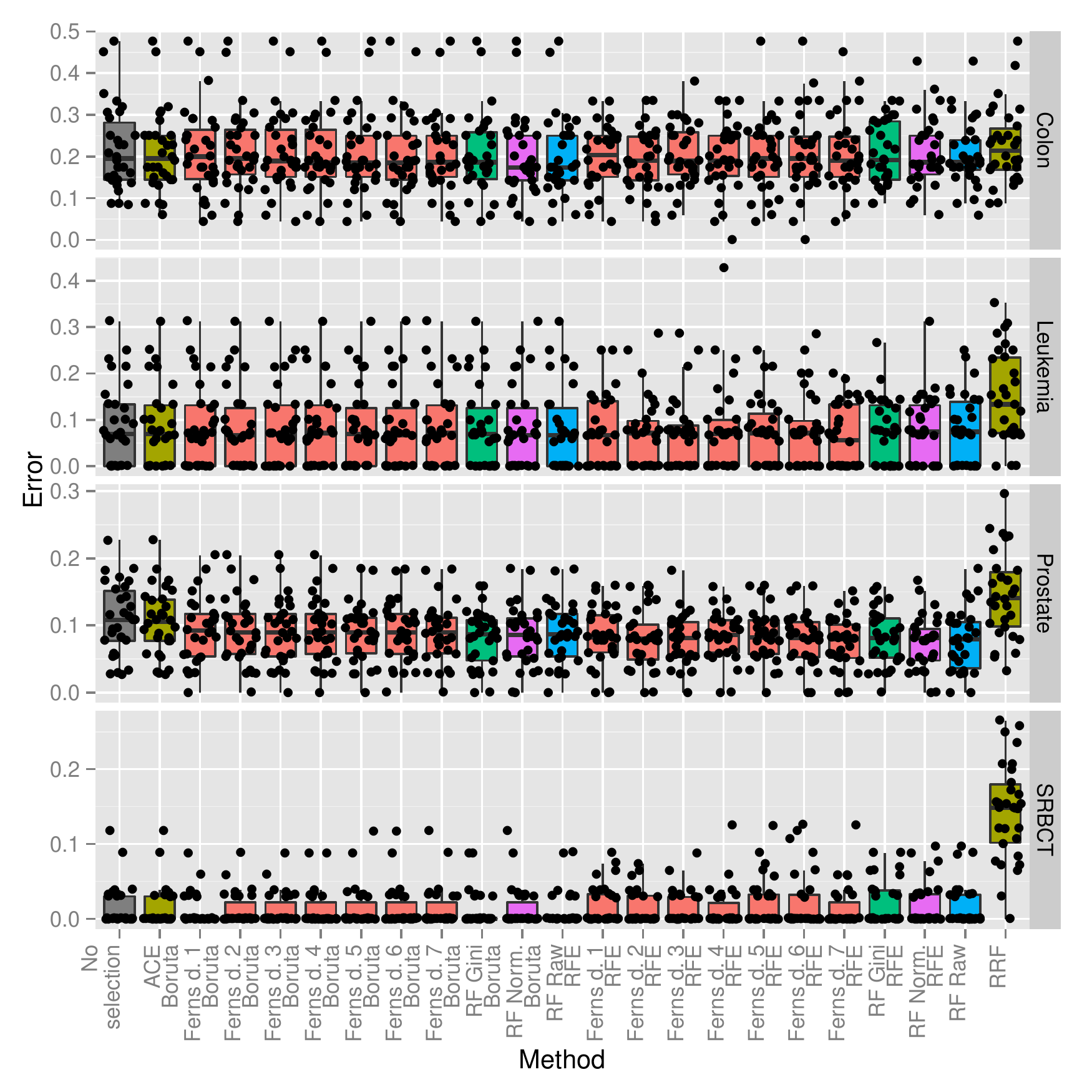}
\caption{\label{fig:err} Post-selection errors of a Random Forest classifier over bootstrap iterations, presented directly and as boxplots. Colour is used for clarity.}
\end{figure}

Following this approach, each set of gene selections sampled in a bootstrap iteration was used to build a corresponding set of Random Forest validation models that were tested on objects not present in the corresponding resamples.
That is, they were not used in feature selection or in the model training step.
These results are presented in Figure~\ref{fig:err}.

It is clear that, with the exception of the RRF method, that all of the other investigated methods produced nearly indistinguishable post-selection errors.
Due to high variability in the results, however, the RRF method produced results for the \Kha{} and \Sin{} sets that were not significantly different from the results of the best performing method in each respective set.
This result also suggests the selection of the Random Ferns' depth and the Random Forest importance source did not influence the post-selection error.

Consequently, although an analysis based on post-selection error will obviously detect the removal of a significant amount of non-redundant information that is usable for the classifier, it is clear that its resolution is too low to serve as a reliable assessment measure of gene selection quality.
Because the post-selection error is also a highly variable statistic, one should never rely on a single estimate of its value.
In the most striking example from this analysis, the application of the RFE method to the \Alo{} dataset produced a range of error values over all iterations sampled that varied by almost 50\% (producing random guesses as well as perfect classification).

On the other hand, no significant improvement over the models built from an entire dataset was observed.
This result demonstrates the established fact that, due to its ensemble construction, the Random Forest method can handle a large number of noisy features without a significant increase in effort.

\subsection{Self-consistency}
Gene selection quality was assessed by comparing the sets of genes selected by a given method over the 30 bootstrap iterations.
From these data, genes that were selected in more iterations of the bootstrap than would be expected to occur at random were identified as significant selections; these genes are referred to as \textit{significantly self-consistent selections} (SCSs) in this paper.

\begin{table}
\begin{tabular}{r||rrr|rrr|rrr|rrr}

Method & \multicolumn{3}{|c|}{\Alo} & \multicolumn{3}{|c|}{\Gol} & \multicolumn{3}{|c|}{\Kha} & \multicolumn{3}{|c}{\Sin} \\
& $c$ & $f$ & $c/f$ & $c$ & $f$ & $c/f$ & $c$ & $f$ & $c/f$ & $c$ & $f$ & $c/f$ \\
\hline
\hline

RF-ACE & 0.0 & 1354.2 & 0\% & 398.0 & 1946.1 & 20\% & 0.0 & 1569.1 & 0\% & 1356.0 & 7778.6 & 17\% \\
\hline
Bor. Ferns  1 & 91.8 & 176.6 & 52\% & 228.9 & 391.9 & 58\% & 336.8 & 567.8 & 59\% & 480.3 & 757.3 & 63\% \\
Bor. Ferns  2 & 93.0 & 182.8 & 51\% & 249.0 & 423.3 & 59\% & 354.5 & 652.2 & 54\% & 520.7 & 840.4 & 62\% \\
Bor. Ferns  3 & 104.9 & 192.2 & 55\% & 247.5 & 439.6 & 56\% & 375.0 & 720.2 & 52\% & 582.1 & 916.6 & 64\% \\
Bor. Ferns  4 & 118.8 & 210.8 & 56\% & 252.9 & 453.0 & 56\% & 383.6 & 786.7 & 49\% & 621.9 & 986.5 & 63\% \\
Bor. Ferns  5 & 120.6 & 227.2 & 53\% & 270.9 & 482.7 & 56\% & 396.2 & 864.2 & 46\% & 670.3 & 1046.3 & 64\% \\
Bor. Ferns  6 & 135.9 & 246.8 & 55\% & 275.3 & 513.2 & 54\% & 395.8 & 959.4 & 41\% & 692.1 & 1077.3 & 64\% \\
Bor. Ferns  7 & 145.0 & 277.9 & 52\% & 296.4 & 550.1 & 54\% & 357.0 & 1058.3 & 34\% & 705.8 & 1104.5 & 64\% \\
Bor. RF Gini & 77.2 & 137.8 & 56\% & 230.2 & 407.6 & 56\% & 358.4 & 626.7 & 57\% & 267.2 & 462.1 & 58\% \\
Bor. RF Raw & 116.9 & 214.7 & 54\% & 256.9 & 446.2 & 58\% & 403.9 & 807.6 & 50\% & 422.7 & 728.0 & 58\% \\
Bor. RF Norm. & 103.3 & 199.1 & 52\% & 237.5 & 403.3 & 59\% & 400.8 & 839.2 & 48\% & 301.5 & 529.9 & 57\% \\
\hline
\hline
RFE Ferns  1 & 23.2 & 95.5 & 24\% & 4.4 & 8.5 & 51\% & 39.0 & 72.8 & 54\% & 28.9 & 503.9 & 6\% \\
RFE Ferns  2 & 18.6 & 55.2 & 34\% & 4.4 & 8.0 & 55\% & 36.6 & 75.2 & 49\% & 73.8 & 854.0 & 9\% \\
RFE Ferns  3 & 23.1 & 88.5 & 26\% & 4.3 & 8.3 & 52\% & 30.6 & 78.1 & 39\% & 47.2 & 125.9 & 38\% \\
RFE Ferns  4 & 18.0 & 77.3 & 23\% & 3.9 & 8.5 & 46\% & 38.6 & 70.9 & 54\% & 34.9 & 402.9 & 9\% \\
RFE Ferns  5 & 18.6 & 52.5 & 35\% & 4.9 & 9.1 & 54\% & 38.0 & 104.3 & 36\% & 99.5 & 321.1 & 31\% \\
RFE Ferns  6 & 18.5 & 58.7 & 32\% & 5.1 & 9.6 & 53\% & 33.1 & 52.5 & 63\% & 75.6 & 280.8 & 27\% \\
RFE Ferns  7 & 13.8 & 70.9 & 19\% & 5.0 & 9.6 & 52\% & 32.8 & 49.1 & 67\% & 36.6 & 81.3 & 45\% \\
RFE RF Gini & 17.7 & 110.1 & 16\% & 4.8 & 8.5 & 57\% & 26.5 & 38.9 & 68\% & 71.7 & 163.2 & 44\% \\
RFE RF Raw & 18.6 & 51.2 & 36\% & 4.8 & 8.3 & 58\% & 31.3 & 46.9 & 67\% & 43.6 & 274.9 & 16\% \\
RFE RF Norm. & 11.9 & 32.5 & 37\% & 4.3 & 8.0 & 53\% & 28.1 & 43.7 & 64\% & 34.6 & 60.0 & 58\% \\
\hline
RRF & 1.4 & 15.9 & 9\% & 0.0 & 3.8 & 0\% & 1.9 & 8.3 & 22\% & 1.1 & 19.2 & 6\% \\

\hline
\hline

No. features & \multicolumn{3}{|c|}{2000} & \multicolumn{3}{|c|}{3051} & \multicolumn{3}{|c|}{1586} & \multicolumn{3}{|c}{12533} \\

\end{tabular}

\caption{\label{tab:cons} The average number of significantly self-consistent and all selected genes by a given method in one bootstrap iteration. $c$ -- the average number of significantly self-consistent genes, $f$ -- the average number of selected genes.}
\end{table}

Table~\ref{tab:cons} summarises the average number of self-consistent and all selected genes as well as their ratios for all investigated sets and methods.
It is clear that the RF-ACE algorithm selected the most genes for all sets, with values ranging from 62\% to 99\% of all present genes in a set.
However, in the case of the \Alo{} and \Kha{} sets, the fraction SCSs was negligible, while in the case of the Leukaemia and Prostate datasets, it reached only approximately 20\%.
These results suggest that this method produces a large number of false positives that overwhelm the signal.

Overall, the highest number of SCSs were produced by the Boruta method; in the best cases, the SCSs covered 56--64\% of all selections and approximately 55\% on average.
While more SCSs were found in all sets using the Random Ferns importance measure than with any of the Random Forest-based measures, the difference was noticeable only in the case of the \Sin{} set.
Moreover, the use of both algorithms led to very similar SCS ratios.
Out of the Random Forest-based importance measures, there was no measure that was clearly the best, but the raw importance measure seemed to be the most reliable choice.
The increase in the Random Ferns depth parameter consistently contributed to an increase in the number of genes found by the Boruta method.
For the \Alo{}, \Gol{} and \Sin{} datasets this effect was accompanied by a proportional increase in the number of SCSs, which caused the SCS ratio to be approximately constant.
This was not the case for the \Kha{} set, however.
In the \Kha{} set, the number of SCSs did not increase and, therefore, its ratio dropped with the fern depth.
Still, the overall performance of the Boruta method was surprisingly stable across the investigated importance sources, and it is unlikely that an incorrect set-up will substantially diminish its performance.

As expected for a minimal-optimal method, the RFE algorithm selected a much smaller number of genes than the RF-ACE or Boruta methods (selecting, on average, from 0.2--3.9\% of all genes in a set).
The number of SCSs was fairly stable in all sets except the \Sin{} set, which produced SCS numbers that were approximately an order of magnitude smaller when the Boruta method was used.
However, the number of found genes varied in an inconsistent manner across different importance sources.
While the SCS ratios in the \Gol{} and \Kha{} sets were reasonably stable and reached 58\% and 73\%, respectively, the SCS ratios were less than 40\% in the \Alo{} set and ranged from 6\% to 58\% in the \Sin{} set (with an average of 28\%).
Therefore, it is likely that the minimal-optimal sets still contained a significant fraction of irrelevant genes (although in much smaller numbers than that produced by the all-relevant methods).
Moreover, RFE's results can be significantly altered by the importance source.

The RRF algorithm selected the least number of genes from all sets, ranging from 4 to 20 (or 0.1\% to 0.8\%, respectively, of all genes in a set).
Moreover, the results from the RRF algorithm were very inconsistent.
The largest significant average selection made by the RRF algorithm was 22\% in the \Kha{} set, while the number of consistent genes found by the RRM algorithm never exceeded 2.

\subsection{Execution time}

\begin{table}
\begin{tabular}{r||r|r|r|r}

Method & \Alo & \Gol & \Kha & \Sin \\
\hline
\hline

RF-ACE &  40' &  24' &  57' & 2h 47' \\
\hline
Boruta Ferns depth 1 &  01' &  01' &  01' &  03' \\
Boruta Ferns depth 7 &  05' &  05' &  11' &  09' \\
Boruta RF Gini & 2h 27' & 2h 19' & 10h 52' & 30h 48' \\
Boruta RF Raw & 3h 30' & 2h 43' & 14h 35' & 40h 23' \\
Boruta RF Norm. & 3h 28' & 2h 34' & 16h 04' & 35h 27' \\
\hline
\hline
RFE Ferns depth 1 &  10' &  08' &  15' & 6h 43' \\
RFE Ferns depth 7 &  10' &  08' &  16' & 7h 24' \\
RFE RF Gini &  21' &  16' &  31' & 13h 34' \\
RFE RF Raw &  21' &  16' &  33' & 13h 49' \\
RFE RF Norm. &  22' &  17' &  32' & 13h 17' \\
\hline
RRF &  03' &  02' &  04' & 1h 04' \\

\hline\hline

No. features & 2000 & 3051 & 1586 & 12533 \\
No. objects & 62 & 38 & 83 & 102 \\

\end{tabular}

\caption{\label{tab:tim} The execution time of selected algorithms, represented as the mean over 30 bootstrap iterations. All algorithms investigated in this study were run single-threaded.}
\end{table}

The average execution time of the selected algorithms is provided in the Table~\ref{tab:tim}.
The slowest method was the Boruta algorithm using the Random Forest importance measure, with computational training time ranged from hours to days, especially for larger sets.
The RF-ACE and RRF algorithms required far less execution time, which never exceeded 1 hour for the \Alo{}, \Gol{} and \Kha{} sets or 2.5 hours for the much larger \Sin{} set.

However, while the difference in the computational time of the Boruta algorithm was minor for Random Forest importance sources, the employment of Random Ferns resulted in significant increases in speed that ranged from 20 to 200 times faster.
Consequently, the execution time of the Boruta algorithm was comparable to or even shorter than that of RF-ACE and RRF.
In the case of RFE, the gain from using Random Ferns was much smaller because this algorithm also relies on Random Forest for assessing the classifier accuracy from the current subset of genes.

\section{Conclusions}
As far as post-selection classification accuracy is concerned, all investigated methods were effectively equivalent.
This proves that assessing gene selection algorithms in this way may be deceiving or inconclusive and, therefore, calls for deep and careful investigation of the significance of the observed accuracy differences.

Out of all the analysed methods, the Boruta algorithm found the most genes predicted to be important and, at the same time, achieved the highest ratio of self-consistent selections in its results.
Although it remains unknown how many of these novel genes are biologically relevant, these results provide strong justification that the selections generated by this method are promising candidates that should be explored further to identify more subtle aspects of the phenomena investigated via microarray experiments.

Despite the fact that Boruta requires an impractical amount of computation time in its default set-up, using the importance source produced by the Random Ferns algorithm decreased its running time to levels comparable with other investigated methods without sacrificing or improving the selection quality.

As expected, the minimal-optimal RFE and RRF methods selected a much smaller subset of genes than the all of the relevant methods.
However, the RFE and RRF methods achieved a similar level of selection stability and, thus, also generated a substantial amount of false positives.
This result suggests that, even when focused on the most pronounced associations, it is important to be aware of the effects of the $p\gg n$ issues that are inherent to microarray data.

\section{Methods}
\subsection*{Feature selection algorithms}
Both the RF-ACE \cite{rface,Tuv2009} and Boruta \cite{Kursa2010c} algorithms are based on the idea first introduced by \cite{Stoppiglia2003}.
That is, they extend the information system with \textit{shadows}, which are artificial features created by permuting the order of values in the original data, and then using shadows' importance scores to judge the significance of the scores obtained by the actual features.

The algorithms differ in the testing scheme used, however.
RF-ACE performs a predefined number of iterations (the default value used in this study is 20) and, at each step, collects the importance of real features and the mean importance of all shadows.
For each feature, Student’s t-test is applied to check whether its mean importance in the iterations is significantly larger than the mean importance of the shadow attributes.
Features with $p$-values less than 0.05 are returned as relevant.

On the other hand, Boruta checks which features in an iteration achieved higher importance than the best shadow; such events are counted for each feature until their number becomes either significantly higher or lower than what is expected at random, using a default $p$-value cut-off of 0.01.
In the first case, the feature is deemed relevant and in the latter case, it is deemed irrelevant, which leads to the removal of the feature and its shadow from the information system.
This procedure is repeated until the status of all features is decided or until a previously set limit of iterations is exhausted, in which case, the status of some features may be undecided.
To make fair comparisons with methods that perform only a relevance test, in this work, all undecided features are assumed to be irrelevant.

Both RF-ACE and Boruta re-shuffle shadow features after every iteration.

Recursive Feature Elimination (RFE) is a group of methods where selection is performed by iterative stripping of less important features from the set until the classifier error becomes minimal.
There are many implementations of this method that differ in the importance of the source used, the stripping criterion and the accuracy assessment method.
In this study, the following algorithm was adapted from the R caret \cite{caret} package.
First, the accuracy is assessed via 10 iterations of bootstrap validation of a 50000-tree Random Forest classifier and stored along with the current list of features.
Then, the arbitrary importance source is applied to a set.
This result is used to remove the least important features so that the number of features will decrease to the highest power of 2 that is lower than the current number.
This procedure is repeated until the number of features drops to 4.
Finally, the list of features for which the error was minimal is returned as the final selection.

Regularised Random Forest (RRF) is a modification of a Random Forest that incorporates regularisation into the tree growing algorithm \cite{Deng2012a,Deng2012}.
Specifically, RRF establishes a penalty for the use of a feature that was not previously used in a current tree construction.
This penalty is proportional to the potential information gain from building a split on this feature, so that only features with significant information that is not redundant with respect to already built splits will be included in the model.
Obviously, this approach leads to a situation where only a subset of all features is actually used in the ensemble.
This subset represents the final result produced when RRF is used as a feature selection algorithm.

\subsection{Importance sources}
For the importance source, I have used the three importance measures produced by the Random Forest \cite{Liaw2002} as well as the importance score produced by the Random Ferns algorithm, which is a variation of the Random Forest.

The first Random Forest importance measure is the overall decrease in node impurity due to splits performed on certain features, which is expressed as the Gini index (RF Gini).
The second measure is calculated in a per-tree manner by finding the difference between a tree’s accuracy on an original out-of-bag (OOB) subset and its version with randomly permuted objects within the analysed feature.
These values are then averaged.
Because this measure was the only one mentioned in the original Random Forest paper \cite{Breiman2001}, I refer to it here as the raw importance (RF Raw).
The third measure is the raw importance normalised by the standard deviation of accuracy differences over the trees (RF Norm).

The Random Ferns \cite{Ozuysal2008} is a simplified variation of the Random Forest algorithm that is an ensemble of \textit{ferns}, which are modified decision trees with a fixed depth (which is a parameter of the algorithm) and that have the same splitting criterion for all splits at the same level.
While a regular classification tree stores the majority classes in its leaves, a fern stores vectors of class probabilities; to this end, ensemble voting is achieved by a maximum a-posteriori rule instead of by selecting the class with the most votes.
The Random Ferns implementation used in this study, rFerns \cite{Kursa2012a}, produces fern splits at random (i.e., based on a randomly selected feature and a randomly selected threshold).

The original Random Ferns does not produce feature importance.
The one used in this study is native to the rFerns implementation, and is similar to the raw importance of Random Forest, except rFerns considers differences in OOB probabilities for a correct class rather than differences in the number of correct votes.
In this work, I have assessed the importance of rFerns independently for fern depths that range from 1 to 7.

While both methods scan the space of features randomly, it is crucial to build ensembles large enough to ensure all features will have an equal chance to participate in the model and generate a stable importance score.

\subsection{Datasets}

\begin{table}
\begin{tabular}{r||r|r|r|r|r}

Dataset & Reference & Genes & Objects & Classification target & Objects per class\\
\hline
\hline
\Alo & Alon et al \cite{Alon1999} & 2000 & 62 & normal/tumor colon tissue & 40:22 \\
\Gol & Golub et al \cite{Golub1999} & 3051 & 38 & ALL/AML leukemia type & 27:11\\
\Kha & Khan et al \cite{Khan2001} & 1586 & 83 & 4 SRBCT types  & 11:29:18:25\\
\Sin & Singh et al \cite{Singh2002} & 12533 & 102 & normal/tumor prostate tissue & 50:52\\
\end{tabular}

\caption{\label{tab:sets} The microarray datasets used in this study.}
\end{table}

The testing of all methods enumerated in the previous sections used four well known microarray datasets obtained from actual experiments.
The summary and characteristics of these data are provided in Table~\ref{tab:sets}.

\subsection{Testing and assessment of the results}
First, to perform the bootstrap estimation, each dataset was used to create 30 resampled sets that were obtained by sampling with replacement an equal number of objects as was present in the original set.

Then, each method of gene selection was executed on all resampled sets, and the results were used to identify SCSs.
First, the expected distribution of the number of selections was estimated from a binomial distribution with parameter $p$ approximated as the mean fraction of the selected features in each iteration.
This distribution was then used to find genes with a number of selections significantly higher than would be expected by random chance with a $p$-value of 0.01.
The Holm–Bonferroni \cite{Holm1979} correction was applied to remove the effect of multiple testing.
These selections were then identified as significantly self-consistent and their count was averaged over all iterations.

Next, all investigated methods were tested by the analysis of post-selection error made by a classifier trained on a set reduced to the selected genes.
For this purpose, for each bootstrap iteration, a Random Forest model composed of 50000 trees was trained on a set reduced to objects belonging to the respective resampled set as well as features that were selected by the given method; then, this model was tested on the remaining objects that were not used in its training.
The obtained predictions were also used to assess the significance of the accuracy differences between methods.
This was accomplished using a paired one-sided Holm-Bonferroni-corrected Mann-Whitney-Wilcoxon test with a $p$-value of 0.01 to compare the errors from each bootstrap iteration of a given method to the errors from the best performing method on a particular dataset.
The use of a non-parametric test was required because of the non-normal distribution of the errors across the iterations.

Finally, the running times of all executed algorithms were collected.
In order to make comparison meaningful, all calculations were performed on a homogeneous cluster of AMD Opteron 835X x86\_64 Linux machines, using R 2.15.0 \cite{R}, randomForest 4.6-6, rFerns 0.3.1, RRF 1.2 and RF-ACE 1.1.0.
Moreover, each algorithm was run single-threaded.

\section*{Acknowledgements}
\small
This work was financed by the National Science Centre, grant 2011/01/N/ST6/07035.
Computations were performed at the ICM UW, grant G48-6.

\bibliographystyle{plain}
  \bibliography{hiperbor}      

\end{document}